\documentclass{llncs}
\usepackage{makeidx}  
\usepackage{amssymb}
\usepackage[cmex10]{amsmath}
\usepackage{graphicx}
\usepackage{float}
\usepackage{subfig}
\usepackage{bm}
\begin{document}
%
%
\title{Semi-supervised Dictionary Learning Based on Hilbert-Schmidt Independence Criterion}
\titlerunning{Semi-supervised Dictionary Learning}  
%
\author{Mehrdad~J.~Gangeh\inst{1} \and Safaa~M.A.~Bedawi\inst{2} \and Ali~Ghodsi\inst{3}
\and Fakhri~Karray\inst{2}}
\authorrunning{Mehrdad~J.~Gangeh et al.}   
%
\tocauthor{Mehrdad J. Gangeh, Safaa M.A. Bedawi, Ali Ghodsi, Fakhri Karray}

\institute{Departments of Medical Biophysics, and Radiation Oncology, \\
University of Toronto, Canada\\
\email{mehrdad.gangeh@utoronto.ca}
\and
Center for Pattern Analysis and Machine Intelligence, \\
Department of Electrical and Computer Engineering, University of Waterloo, Canada\\
\email{\{sbedawi,karray\}@uwaterloo.ca}
\and
Department of Statistics and Actuarial Science, University of Waterloo, Canada\\
\email{aghodsib@uwaterloo.ca}
}

\maketitle              

\begin{abstract}
In this paper, a novel semi-supervised dictionary learning and sparse representation (SS-DLSR) is proposed. The proposed method benefits from the supervisory information by learning the dictionary in a space where the dependency between the data and class labels is maximized. This maximization is performed using Hilbert-Schmidt independence criterion (HSIC). On the other hand, the global distribution of the underlying manifolds were learned from the unlabeled data by minimizing the distances between the unlabeled data and the corresponding nearest labeled data in the space of the dictionary learned. The proposed SS-DLSR algorithm has closed-form solutions for both the dictionary and sparse coefficients, and therefore does not have to learn the two  iteratively and alternately as is common in the literature of the DLSR. This makes the solution for the proposed algorithm very fast. The experiments confirm the improvement in classification performance on benchmark datasets by including the information from both labeled and unlabeled data, particularly when there are many unlabeled data.
\end{abstract}
\section{Introduction}
\label{sec:intro}

Dictionary learning and sparse representation (DLSR) is one of the most successful mathematical models, which has led to state-of-the-art results in various applications such as face recognition~\cite{DL:Zhong07,DL:Wright09,DL:MengYang11}, image denoising~\cite{DL:Mairal08}, texture classification~\cite{Gangeh:ICIAR11}, and emotion recognition~\cite{Gangeh:TASLP14}. DLSR, however, was originally proposed in an unsupervised setting~\cite{book:Elad}. The main objective function in the optimization problem related to DLSR is to minimize the reconstruction error between the original signal and the reconstructed one in the space of learned dictionary without including the information on class labels into the learning process. To formally describe the original DLSR formulation, we suppose that there is a finite set of data samples denoted as $\mathbf{X}=[\mathbf{x}_1,...,\mathbf{x}_n]\in\mathbb{R}^{d\times n}$, where $d$ is the dimensionality of the data and $n$ is the number of data samples. In original DLSR, the data is decomposed using a few dictionary atoms by optimizing the empirical cost function
\begin{equation}\label{eq:loss1}
L(\mathbf{X},\mathbf{D},\bm{\alpha})=\sum_{i=1}^{n}l(\mathbf{x}_i,\mathbf{D},\bm{\alpha}),
\end{equation}
where $\mathbf{D}\in\mathbb{R}^{d\times k}$ is a dictionary of $k$ atoms, $\bm{\alpha}\in\mathbb{R}^{k\times n}$ are the sparse coefficients and $L, l$ are loss functions. In the literature of the DLSR, the reconstruction error, in mean-squared sense, between the original signal and the reconstructed signal is the most common loss function, which is usually regularized by the $\ell_1$ norm to induce sparsity into the coefficients. Thus, the formulation in~(\ref{eq:loss1}) can be written as
\begin{equation}\label{eq:lasso}
L(\mathbf{X},\mathbf{D},\bm{\alpha})=\min_{\mathbf{D},\bm{\alpha}}\sum_{i=1}^{n}\begin{pmatrix}\frac{1}{2}\|\mathbf{x}_i-\mathbf{D}\bm{\alpha}_i\|_{\textup{2}}^{2}+\lambda\|\bm{\alpha}_i\|_{1}\end{pmatrix},
\end{equation}
where $\bm{\alpha_i}$ is the \emph{i}th column of $\alpha$. In order to avoid arbitrarily large values for $\mathbf{D}$ and consequently, arbitrarily small values for $\alpha$, we need an additional constraint on the dictionary atoms to limit their $\ell_2$ norm to be smaller than or equal to one. The complete optimization problem in~(\ref{eq:lasso}) after adding this constraint is as follows:
\begin{equation}\label{eq:lassoConst}
\begin{aligned}
L(\mathbf{X},\mathbf{D},\bm{\alpha})=
& \min_{\mathbf{D},\bm{\alpha}}\
& & \sum_{i=1}^{n}\begin{pmatrix}\frac{1}{2}\|\mathbf{x}_i-\mathbf{D}\bm{\alpha}_i\|_{\textup{2}}^{2}+\lambda\|\bm{\alpha}_i\|_{1}\end{pmatrix}, \\
& \text{s.t.}
& & \|\mathbf{d}_j\|_2^2\leq1 \;\;\;\;   \forall j=1,..,,k.
\end{aligned}
\end{equation}

The original DLSR formulation given in~(\ref{eq:lassoConst}) is unsupervised as the category information has not been taken into consideration in the optimization problem. However, in a supervised learning paradigm, where the ultimate goal is the classification of the data, this setting may not lead to an optimal discriminative dictionary nor coefficients. A more recent attempt in the literature was to incorporate the class labels into the learning of the dictionary and/or coefficients (refer to~\cite{Gangeh:arxiv15} for a review). This modification resulted in a new category of DLSR, namely called supervised dictionary learning and sparse representation (S-DLSR). Improvements (some significant) over unsupervised DLSR have been reported in the literature for the classification tasks~\cite{DL:MengYang11,DL:Mairal08a,DL:Wright10,Gangeh:TSP13}.

Although S-DLSR benefits from the side information available from category information to learn a more discriminative dictionary, unfortunately, gathering labeled data is often very expensive and time consuming. Most data available is unlabeled and the sample size of the labeled data is often very small, which has a hindering effect on the discriminative quality of the learned dictionary. Semi-supervised learning (SSL) methods can potentially boost the performance of a machine learning system by utilizing both supervisory information and global data distribution. Using a large amount of unlabeled data, which is usually easily accessible, can improve revealing the manifold global distribution~\cite{SSL:Zhou04}, and compensate for the small sample size of labeled data~\cite{book:Chapelle}.

In this paper, a semi-supervised dictionary learning and sparse representation (SS-DLSR) based on Hilbert-Schmidt independence criterion (HSIC) is proposed. The proposed SS-DLSR approach finds a dictionary based on two criteria: first, the maximization of the dependency between the labeled data and the corresponding category information, and second, minimization of the distances between the unlabeled data and their nearest labeled data. The first criterion guarantees finding the space of maximum discrimination based on the information in the category information and labeled data, whereas the second criterion, guarantees that the unlabeled data remain as close as possible to their nearest-neighbor labeled data. Therefore, the learned dictionary (the projection directions computed by using the aforementioned criteria) benefits from the discriminative power of the category information in the labeled data and proximity information of the unlabeled data as an indication of global manifold distribution. the sparse coefficients are subsequently computed in the space of learned dictionary using the formulation given in~(\ref{eq:lassoConst}).

\section{Semi-supervised Dictionary Learning and Sparse Representation}
\label{sec:SSDL}

\subsection{Problem Statement}\label{subsec:problemStatement}

Let $\mathbf{X}=[\mathbf{x}_1,...,\mathbf{x}_n]\in\mathbb{R}^{d\times n}$ be $n$ data samples with the dimensionality of $d$. There are $n_l$ labeled and $n_u$ unlabeled data samples, where $n=n_l+n_u$. Let $\{(\mathbf{x}_1,\mathbf{y}_1),...,(\mathbf{x}_{n_l},\mathbf{y}_{n_l})\}$ be the pair of labeled data ($\mathbf{X}_l\in\mathbb{R}^{d\times n_l}$) and the corresponding labels ($\mathbf{Y}\in\{0,1\}^{c\times n_l}$, where $c$ is the number of classes), and $\mathbf{X}_u=[\mathbf{x}_{n_l+1},...,\mathbf{x}_n]\in\mathbb{R}^{d\times n_u}$ be the unlabeled data samples. We would like to find a dictionary, which can be considered as a transformation, based on two criteria (1) maximizing the dependency between the labeled data $\mathbf{X}_l$ and the labels $\mathbf{Y}$, and (2) minimizing the distance between each unlabeled data with the nearest label data. The first criterion is to guarantee finding a discriminative dictionary using the labeled data, and the second criterion is to ensure the unlabeled data samples are mapped close to their neighboring labeled data and therefore, the global connectivity of data is maintained in the space of the learned dictionary.

The first criterion is implemented using the Hilbert-Schmidt independence criterion (HSIC), which will be explained in the next subsection followed by the design of the dictionary and sparse coefficients for the proposed semi-supervised method.

\subsection{Hilbert-Schmidt Independence Criterion}
\label{ssec:HSIC}

HSIC is a kernel-based measure of independence between two random variables $\mathcal{X}$ and $\mathcal{Y}$ proposed first by Gretton \emph{et al.}~\cite{HSIC:Gretton05a,HSIC:Gretton05b}. It is computed based on the Hilbert-Schmidt norm of cross covariance operators in reproducing kernel Hilbert spaces (RKHSs)~\cite{HSIC:Gretton05b}.

Our focus here is the empirical HSIC, which is computed using a finite set of data samples. To this end, considering $\mathcal{Z}:=\{(\mathbf{x}_1,\mathbf{y}_1,),...,(\mathbf{x}_{n_l},\mathbf{y}_{n_l})\}\subseteq\mathcal{X}\times\mathcal{Y}$ as $n_l$ independent observations drawn from joint probability distribution $P_{\mathcal{X}\times \mathcal{Y}}$, the empirical HSIC is computed using
\begin{equation}\label{eq:EmpHSIC}
\textup{HSIC}(\mathcal{Z})=\frac{1}{(n_l-1)^2}\textup{tr}(\mathbf{KHBH}),
\end{equation}
where $\textup{tr}$ is the trace operator, and $\mathbf{K}$, $\mathbf{B}$, $\mathbf{H}\in\mathbb{R}^{n_l\times n_l}$. $\mathbf{K}$ and $\mathbf{B}$ are kernels on the data and labels, respectively. $\mathbf{H}=\mathbf{I}-n_l^{-1}\mathbf{ee}^{\top}$, where $\mathbf{I}$ is an identity matrix, $\mathbf{e}$ is a vector of all ones and therefore, $\mathbf{H}$ is a centering matrix. Since the empirical HSIC given in~(\ref{eq:EmpHSIC}) is a measure of dependency between $\mathcal{X}$ and $\mathcal{Y}$, in order to maximize this dependency, $\textup{tr}(\mathbf{KHBH})$ should be maximized.

\subsection{Dictionary Learning}
\label{ssec:DictionaryLearning}

As mentioned in the problem statement (Subsection~\ref{subsec:problemStatement}), the dictionary is learned based on two criteria. In order to maximize the dependency between the labeled data and the corresponding labels, as shown in~\cite{Gangeh:TSP13}, the following optimization problem has to be solved:
\begin{equation}\label{eq:SDLHSICConst}
\begin{aligned}
& \underset{\mathbf{D}}{\text{max}}
& & \text{tr}(\mathbf{D}^{\top}\mathbf{X}_l\mathbf{HBH}\mathbf{X}_l^{\top}\mathbf{D}), \\
& \text{s.t.}
& & \mathbf{D}^{\top}\mathbf{D}=\mathbf{I}
\end{aligned}
\end{equation}
where $\mathbf{H}$ is the centering matrix, $\mathbf{B}$ is a kernel on labels, and $\mathbf{D}$ is the dictionary to be learned.
By a few manipulations on the objective function given in~(\ref{eq:SDLHSICConst}), it can be demonstrated that it is another form of empirical HSIC:
\begin{align}\label{eq:SDLToHSIC}    
  \makebox[2em][l]{$\underset{\mathbf{D}}{\text{max}}\; \text{tr}(\mathbf{D}^{\top}\mathbf{X}_l\mathbf{HBH}\mathbf{X}_l^{\top}\mathbf{D})$}  \nonumber \\
  &=\underset{\mathbf{D}}{\text{max}}\; \text{tr}(\mathbf{X}_l^{\top}\mathbf{D}\mathbf{D}^{\top}\mathbf{X}_l\mathbf{HBH}) \nonumber \\
  &=\underset{\mathbf{D}}{\text{max}}\; \text{tr}\bigg(\bigg[(\mathbf{D}^{\top}\mathbf{X}_l)^{\top}\mathbf{D}^{\top}\mathbf{X}_l\bigg]\mathbf{HBH}\bigg) \nonumber \\
  &=\underset{\mathbf{D}}{\text{max}}\; \text{tr}(\mathbf{KHBH}),
\end{align}
where $\mathbf{K}=(\mathbf{D}^{\top}\mathbf{X}_l)^{\top}\mathbf{D}^{\top}\mathbf{X}_l$ is a linear kernel on the projected labeled data into the space of learned dictionary $\mathbf{D}$. As can be clearly observed from the last statement in~(\ref{eq:SDLToHSIC}), the objective function in~(\ref{eq:SDLHSICConst}) has the form of the empirical HSIC and thus, the dictionary $\mathbf{D}$ projects the labeled data to the space of maximum dependency with the corresponding labels.

The second criterion is to minimize the distances between the unlabeled data and the nearest neighbor labeled data in the space of the dictionary learned. In other words, considering $\mathbf{z}=\mathbf{D}^{\top}\mathbf{x}$ as a projected data sample to the space of the learned dictionary, we would like to:
\begin{equation}\label{eq:proximity}
\underset{\mathbf{D}}{\text{min}}
\frac{1}{2}\sum_{i=1}^{n_l}\sum_{j=1}^{n_u}w_{i,j}(\mathbf{z}_i - \mathbf{z}_j)^2, \\
\end{equation}
where $w_{i,j}$ are the weights that define the proximity (neighborhood) of the unlabeled to labeled data. One way to define it is based one nearest neighbor, i.e., $w_{i,j}=1$ if the $j$th unlabeled data is the nearest to the $i$th labeled data and $w_{i,j}=0$ otherwise.

It can be shown~\cite{misc:Luxburg07} that the objective function given in~(\ref{eq:proximity}) can be written in matrix form as follows:
\begin{equation}\label{eq:proximityMatrix}
\underset{\mathbf{D}}{\text{min}}
\frac{1}{2}\sum_{i=1}^{n_l}\sum_{j=1}^{n_u}w_{i,j}(\mathbf{z}_i - \mathbf{z}_j)^2=\underset{\mathbf{D}}{\text{min}}\;\;\;\textup{tr}(\mathbf{ZLZ}^{\top})=\underset{\mathbf{D}}{\text{min}}\;\;\;\textup{tr}(\mathbf{D}^{\top}\mathbf{XLX}^{\top}\mathbf{D}), \\
\end{equation}
where $\mathbf{L}$ is the Laplacian of the graph made by the projected data points $\mathbf{Z}=[\mathbf{z}_1,...,\mathbf{z}_n]$ in the space of learned dictionary, and is defined as $\mathbf{L}=\mathbf{Q}-\mathbf{W}$, where $\mathbf{W}(i,j)=w_{i,j}$ and $\mathbf{Q}$ is a diagonal matrix, where $q_{i,i}=\sum_j w_{i,j}$.

Combining the two objective functions given in~(\ref{eq:SDLHSICConst}) and~\ref{eq:proximityMatrix}, the overall optimization problem for the computation of the dictionary can be written as follows:

\begin{equation}\label{eq:SSDL}
\begin{aligned}
& \underset{\mathbf{D}}{\text{max}}
& & \text{tr}\begin{bmatrix}\mathbf{D}^{\top}\begin{pmatrix}(1-\eta)\mathbf{X}_l\mathbf{HBH}\mathbf{X}_l^{\top} - \eta\;\mathbf{XLX}^{\top}\end{pmatrix}\mathbf{D}\end{bmatrix}, \\
& \text{s.t.}
& & \mathbf{D}^{\top}\mathbf{D}=\mathbf{I}
\end{aligned}
\end{equation}
where $0\leq\eta\leq 1$ is a constant that determines the relative contributions of the two terms in the objective function. According to the Rayleigh-Ritz theorem~\cite{book:Lutkepohl96}, the solution for the optimization problem given in~(\ref{eq:SSDL}) is the corresponding eigenvectors of the largest eigenvalues of $\mathbf{\Phi}=(1-\eta)\mathbf{X}_l\mathbf{HBH}\mathbf{X}_l^{\top} - \eta\;\mathbf{XLX}^{\top}$.

\subsection{Sparse Coefficients}
\label{ssec:sparseCoeff}

After the computation of the dictionary using~(\ref{eq:SSDL}), the sparse coefficients can be computed using the formulation provided in~(\ref{eq:lasso}), which is called \emph{lasso} if the dictionary is known~\cite{DL:Tibshirani96}. Although~(\ref{eq:lasso}) can be solved using fast iterative methods, since the dictionary is orthogonal, as shown in~\cite{DL:Donoho95,DL:Friedman07}, the sparse coefficients can be computed using soft-thresholding with the soft-thresholding operator $S_{\lambda}(.)$:
\begin{equation}\label{eq:softThresholding}
\alpha_{ij}=S_{\lambda}\begin{pmatrix}[\mathbf{D}^{\top}\mathbf{x}_i]_j\end{pmatrix},
\end{equation}
where $\alpha_{i,j}$ is the $(i,j)$th element of $\bm{\alpha}$ and $S_{\lambda}(t)$ is defined as follows:
\begin{equation}
S_{\lambda}(t)=\left\{\begin{matrix}
t-0.5\lambda\;\;\; \textup{if}\;\;t>0.5\lambda \\ \;\;\:t+0.5\lambda\;\;\; \textup{if}\;\;t<-0.5\lambda
\\
0 \;\;\;\;\;\;\;\;\;\;\;\textup{otherwise}
\end{matrix}\right.
\end{equation}

\section{Experiments and Results}
\label{sec:Experiment}

To validate the proposed semi-supervised dictionary learning and sparse representation method (SS-DLSR), two benchmark datasets publicly available from UCI machine learning repository\footnote[1]{http://archive.ics.uci.edu/ml/} were used. The two datasets were the Sonar ($n=208$, $d=60$, and $c=2$) and the Parkinsons ($n=297$, $d=13$, and $c=2$) datasets.

The performance of the proposed SS-DLSR was evaluated for a fixed dictionary size ($k=8$) and varying relative ratio of the labeled to unlabeled data $n_l/(n_l+n_u)$. To this end, 70\% of the data was randomly selected as the training set and 30\% as the test set. The training data was further divided to different ratios of labeled and unlabeled data as shown in Table~\ref{tab:labUnlabRatioExp} ($n_l/(n_l+n_u)=\{0.05, 0.1, 0.3, 0.5\}$). One nearest neighbor was used as the proximity measure between the unlabeled and labeled data to determine the matrix of weights in~(\ref{eq:proximity}). The value of $\eta$ for the computation of the dictionary in~(\ref{eq:SSDL}) was set to three different values, i.e., 0 (ignoring unlabeled data), 1 (ignoring labeled data), and $\eta^*$ (the most discriminative dictionary corresponding to best classification performance). The sparse coefficients were computed for the labeled portion of the training data as well as for the test data. A support vector machine (SVM) with radial basis function (RBF) kernel was used for the classification of the data by submission of the sparse coefficients to the classifier as suggested in~\cite{DL:Raina07}. The SVM was tuned using 5-fold cross validation on the labeled portion of training data to find the optimal kernel width ($\gamma^*$) and optimal trade-off parameter ($C^*$). Subsequently, the SVM was trained on whole labeled data in the training set using the optimal $\gamma^*$ and $C^*$ values and tested on the test set. The experiments were repeated 10 times for different random split of the data to training and test sets. The performance is reported in terms of classifier accuracy (averaged over 10 runs) in Table~\ref{tab:labUnlabRatioExp}.

From the results provided in Table~\ref{tab:labUnlabRatioExp}, there are several immediate observations. First, by adding unlabeled data to the learning of the dictionary (the columns in Table~\ref{tab:labUnlabRatioExp} corresponding with $\eta^*$), the classification performance is increased, which means that the learned dictionary is more discriminative. This reveals that the proposed algorithm can effectively incorporate the information from both labeled and unlabeled data into the learning of the dictionary. Second, by decreasing the rate of labeled to unlabeled data ($n_l/(n_l+n_u)$), the gain in performance from adding unlabeled data is increased. In realistic settings, there usually exist many unlabeled data and only a small number of labeled data. The proposed SS-DLSR algorithm benefits more from the information provided by the unlabeled data in these situations as can be observed by comparing the column corresponding with $\eta^*$ (including both the labeled and unlabeled data into the dictionary learning) and the column with $\eta=0$ (including only labeled data into the dictionary learning).

\begin{table}[t!]
    \caption{The classification rate (\%) of the proposed SS-DLSR algorithm on two benchmark datasets. The results were compared for various settings in the proposed algorithm including different relative contributions of the labeled and unlabeled data on dictionary learning (varying $\eta$), and different ratios of labeled to unlabeled data (varying $n_l/(n_l+n_u)$).}
  \centering
  \label{tab:labUnlabRatioExp}
  \begin{tabular}{|c@{\hskip 0.25in} |c@{\hskip 0.25in} c@{\hskip 0.25in} c@{\hskip 0.25in} |c@{\hskip 0.25in} c@{\hskip 0.25in} c@{\hskip 0.25in} |}
        \hline
        $\frac{n_l}{n_l+n_u}$           & \multicolumn{3}{c|}{Sonar}  & \multicolumn{3}{c|}{Ionosphere}                                              \\  \cline{2-4}  \cline{5-7}
        ~                  & $\eta=0$ & $\eta=1$ & $\eta^*$ & $\eta=0$ & $\eta=1$ & $\eta^*$ \\ \hline \hline
        0.5           & 69.03       & 51.29       & 70.97       & 88.45 & 76.90 & 86.72    \\
        ~             & $\pm$5.78  & $\pm$6.62  & $\pm$3.95  & $\pm$2.82 & $\pm$5.15 & $\pm$3.90       \\
        0.3           & 66.45       & 51.94       & 68.71       & 85.34 & 74.83 & 85.69    \\
        ~             & $\pm$7.40  & $\pm$4.01  & $\pm$8.54  & $\pm$3.92 & $\pm$4.16 & $\pm$3.99     \\
        0.1           & 57.74       & 49.19       & 61.45       & 78.94 & 73.45 & 80.34    \\
        ~             & $\pm$4.97  & $\pm$5.55  & $\pm$8.13  & $\pm$5.50 & $\pm$5.28 & $\pm$6.81     \\
        0.05          & 53.55       & 50.97       & 55.65       & 72.07 & 68.45 & 74.66    \\
        ~             & $\pm$5.98  & $\pm$5.44  & $\pm$8.20  & $\pm$13.52 & $\pm$13.28 & $\pm$6.56     \\
        \hline
    \end{tabular}
\end{table}

\section{Discussion and Conclusion}
\label{sec:disc}

In this paper, a novel semi-supervised dictionary learning and sparse representation method was proposed. A discriminative dictionary was learned in the space of maximum dependency between the labeled data and class labels, where the connectivity of the data was maintained by minimizing the distances between the unlabeled data and the corresponding nearest labeled data. As can be seen from~(\ref{eq:SSDL}), the dictionary has a closed form solution. Also, by using soft-thresholding, the sparse coefficients can be computed using a closed-form solution as given in~(\ref{eq:softThresholding}). The proposed SS-DLSR approach is, therefore, very fast. The effectiveness of the proposed method in learning from both supervisory information (based on labeled data) and graph connectivity information (based on unlabeled data) was demonstrated by experiments on two benchmark datasets from UCI machine learning repository.

\subsubsection{Acknowledgment.}

The first author gratefully acknowledges the funding from the Natural Sciences and Engineering Research Council (NSERC) of Canada under Postdoctoral Fellowship (PDF-454649-2014).

%
%

\bibliographystyle{splncs}
\bibliography{D:/Mehrdad/Bib_Centeralized/refNCD,D:/Mehrdad/Bib_Centeralized/refTexture}

\end{document}